\documentclass[sn-mathphys,Numbered,iicol]{sn-jnl}

\usepackage{graphicx}%
\usepackage{multirow}%
\usepackage{amsmath,amssymb,amsfonts}%
\usepackage{amsthm}%
\usepackage{mathrsfs}%
\usepackage[title]{appendix}%
\usepackage{xcolor}%
\usepackage{textcomp}%
\usepackage{manyfoot}%
\usepackage{booktabs}%
\usepackage{algorithm}%
\usepackage{algorithmicx}%
\usepackage{algpseudocode}%
\usepackage{listings}%

\usepackage{mathtools}
\usepackage{cuted}
\usepackage{hyperref}

\theoremstyle{thmstyleone}%
\theoremstyle{thmstyletwo}%
\newtheorem{remark}{Remark}%

\theoremstyle{thmstylethree}%

\begin{document}

\title[Parkinson gait modelling from an anomaly deep representation]{\textbf{Parkinson gait modelling from an anomaly deep representation}}


\author[1]{\fnm{Edgar} \sur{Rangel}}\email{edgar.rangel@correo.uis.edu.co}

\author*[1]{\fnm{Fabio} \sur{Mart\'inez}}\email{famarcar@saber.uis.edu.co}

\affil*[1]{\orgdiv{Biomedical Imaging, Vision and Learning Laboratory (BIVL$^2$ab)}, \orgname{Universidad Industrial de Santander}, \orgaddress{\city{Bucaramanga}, \state{Santander}, \country{Colombia}}}


\abstract{Parkinson's Disease (PD) is associated with gait movement disorders, such as bradykinesia, stiffness, tremors and postural instability, caused by progressive dopamine deficiency. Today, some approaches have implemented learning representations to quantify kinematic patterns during locomotion, supporting clinical procedures such as diagnosis and treatment planning. These approaches assumes a large amount of stratified and labeled data to optimize discriminative representations. Nonetheless these considerations may restrict the approaches to be operable in real scenarios during clinical practice. This work introduces a self-supervised generative representation to learn gait-motion-related patterns, under the pretext of video reconstruction and an anomaly detection framework. This architecture is trained following a one-class weakly supervised learning to avoid inter-class variance and approach the multiple relationships that represent locomotion. The proposed approach was validated with 14 PD patients and 23 control subjects, and trained with the control population only, achieving an AUC of 95\%, homocedasticity level of 70\% and shapeness level of 70\% in the classification task considering its generalization.}

\keywords{Anomaly detection, Deep Learning, Semi Supervised, Parkinson Disease}



\maketitle

\section{Introduction}\label{sec:intro}
Parkinson's Disease (PD) is the second most common neurodegenerative disorder, affecting more than 6.2 million people worldwide \cite{vos2017global, dorsey2018parkinson}. PD is characterized by the progressive loss of dopamine, a neurotransmitter involved in the execution of voluntary movements. For this reason, one of the main diagnostic support is based on the observation and analysis of progressive motor disorders, such as tremor, rigidity, slowness of movement (bradykinesia), postural instability, among many other related symptoms \cite{balestrino2020parkinson}. The symptoms and their appear order may vary between patients, being a challenge to determine a definitive and universal biomarker to characterize, diagnose, and follow the patient disease progression.

Particularly, the gait is a multi-factorial and complex locomotion process that involves several subsystems being a comprehensive method to view and evaluate the patient disease progression, in other words, the gait is a holistic review of each involved body part to quantify relevant patterns to diagnose PD. In order to capture the associated kinematics two standard alternatives are used: marker-based or markerless capture setups. In the first case, the locomotion variables are meticulously and accurately recovered by wearable or ground-static sensors that measures important variables during the gait, but this process results in restrictive, intrusive, or alters the natural postural gestures for PD description. On the contrary markerless video strategies (\textit{a.k.a} vision-based) address the marker-based problems but an inherent complexity is gained due to the less accurate measure of gait variables altogether with the increase of error sources during the capture of a sample (\textit{e.g.} the illumination changes or the scenery heterogeneity) \cite{kour2019computer}.

Current works have emerged as key solutions to support the PD characterization and classification from other diseases using both standard alternatives \cite{balaji2021data, kleanthous2020new, alharthi2020gait, chavez2022vision, guayacan2021visualising, hu2019graph}. These approaches have been successful in controlled studies but strongly require a stratified, balanced, and ``well-labeled" dataset to avoid overfitting. Besides, the physicians' experience in the labelling process attach a bias determining the disease and restricts the quantification of standard scales indexes \cite{litjens2017survey}. Even worst, these methodologies are focused in solve classification tasks but remains limited in forecast the performance over different populations of patients and a poor exploration of the model captured patterns to explain the discrimination criteria.

Commonly in real world scenarios having unbalance data is natural, for instance Parkinson represents 2-3\% of the population over 65 years old worldwide \cite{poewe2017parkinson} which indicates a challenge when trying to modelling PD against control patients under balanced conditions. Hence, the anomaly learning is intended to address this by modelling PD as abnormalities when compared to "normal" gait patterns from control population. However, anomaly learning field have been poorly explored as far as we know, due to the prevalence of discriminative tasks, the scarce of public datasets and balanced-forced samples used in the above methods.

Contrary to discriminative field, this works explore the generative tasks with anomaly learning. Following a semi-supervised methodology, a deep 3D net is self-trained under a gait video reconstruction pretext task. The model encode a patient's video into a compact embedding representation containing dynamic gait relationships that allows the discrimination between different classes of patients, when only control population is used for training. The main contributions of this work are summarized as follows:

\begin{itemize}
    \item A 3D Convolutional GAN net dedicated to learn spatio-temporal patterns of gait video-sequences. This architecture coded a new digital biomarker as an embedding vector with the capability to represent hidden kinematic relationships of control population to discriminate againts Parkinson patients.
    
    \item A deep explainable representation which explain the discrimination criteria used by the network. This representation is similar to the gradcam and highlight the anomaly patterns detected by the model on the patient's video.
    
    \item A statistical test framework to validate the capability of the approach in terms of generalization, coverage of data and discrimination capability for any class with different groups between them, \textit{i.e.} evaluate the generalization of Parkinsonian patients, at different stages of the disease, \textit{w.r.t} a control population.
    
\end{itemize}

\section{Current Work}\label{sec:related}

Deep discriminative learning is nowadays the standard methodology in much of the computer vision challenges, demonstrating remarkable results in very different domains. For instance, the Parkinson characterization is achieved from sensor-based and vision-based approaches, following a supervised scheme to capture main observed relationships and to generate a particular predictions about the condition of the patients \cite{kour2019computer}. These approaches in general are dedicated to classifying a control population and patients with Parkinson condition. Gait abnormalities have been of particular interest in Parkinson characterization, even to support early diagnosis \cite{poewe2017parkinson, balestrino2020parkinson}. The sensor-based approaches capture precision kinematics from motion signals, approximating to PD classification, but in many of the cases results invasive, alter natural gestures, and only have recognition capabilities in advanced stages of the disease \cite{guo2022detection}. Other marker based strategies have been dedicated to analyze and discriminate Freezing Of Gait (FOG) using force sensors in both feet \cite{kleanthous2020new}. Contrary, the vision-based approaches exploit postural and dynamic features, from video recordings, but the representations underlies on supervised schemes that requires a large amount of labeled data to learn the inter and intra variability among classes \cite{sun2018convolutional, hu2019graph, guayacan2021visualising, chavez2022vision}.  In these methodologies, Hu \textit{et. al.} focused on study FOG, exploring frontal observations during the walking \cite{hu2019graph}. These learning methodologies also require that training data have well-balanced conditions among classes, \textit{i.e.,} to have the same proportion of sample observations for each of the considered class \cite{kour2019computer, di2020gait}.

Unsupervised, semi-supervised and weakly supervised approaches have emerged as a key alternative to model biomedical problems, with significative variability among observations but limited training samples. However, to the best of our knowledge, these learning methods have been poorly explored and exploited in Parkinson characterization, with some preliminary alternatives that use principles of Minimum Distance Classifiers and K-means Clustering \cite{cho2009vision, chen2011quantification, nomm2016alternative, soltaninejad2018body, kour2019computer, schmarje2021survey}. In such sense, the PD modelling from non-supervised perspective may be addressed from reconstruction and generative tasks \cite{kiran2018overview, chalapathy2019deep}, that help to determine sample distributions and future postural and kinematic events. In fact, the PD patterns distribution modelling results key to understand multi-factorial nature of PD, being determinant to define variations such as laterality affectation of disease, abnormality sources, but also to define patient prognosis, emulating the development of a particular patient during the gait.

\begin{figure*}[h!]
    \centering
    \includegraphics[width=0.9\textwidth]{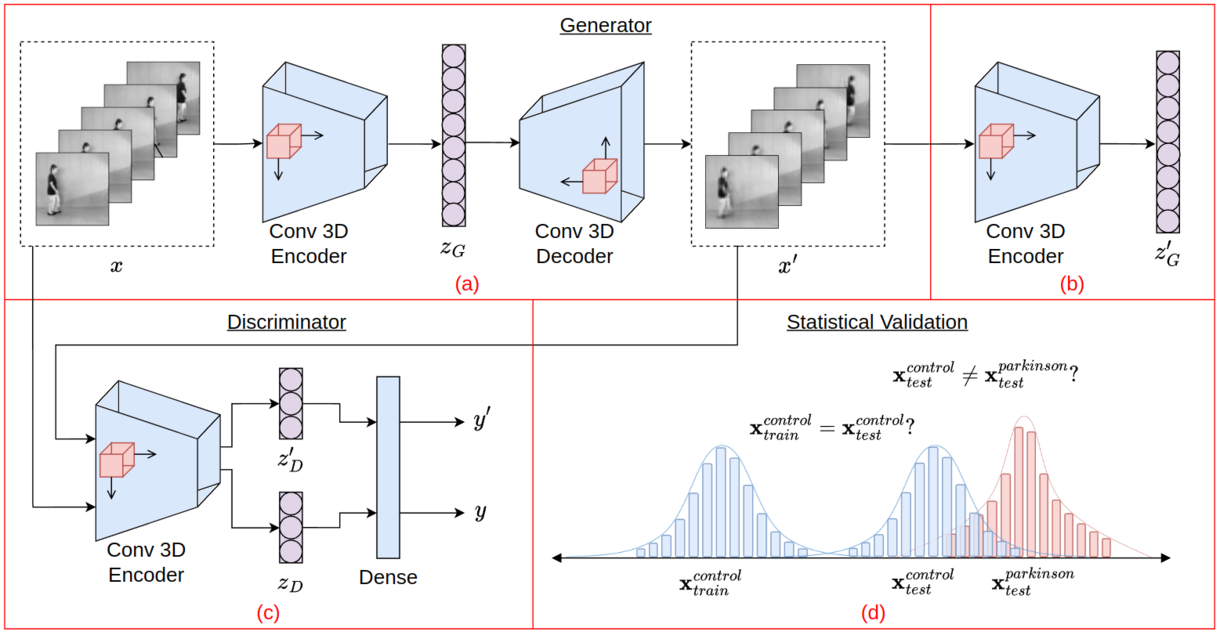}
    \caption{Pipeline of the proposed model separated in volumetric auto-encoder to recover gait patterns (a), Digital gait biomarker (b), Auxiliary task to discriminate reconstructions (c), and statistical validation of learned classes distributions (d)}
    \label{fig:model_arch}
\end{figure*}

\section{Proposed approach}\label{sec:approach}

This work introduces a digital PD biomarker that embedded gait motor patterns, from anomaly video reconstruction task. Contrary to typical classification modeling, we are dedicated to deal with one class learning, \textit{i.e.,} only to learn control gait patterns, approaching the high variability on training samples, without using explicit disease labels. Hence, we hypothesize that a digital biomarker of the disease can be modeled as a mixture of features, composed of samples that were labeled as outliers, from learned representation. In consequence, we analyze the embedding, reconstruction, and discrimination space to later define rules to separate Parkinson from control vectors, during test validation. The general pipeline of the proposed approach is illustrated in Figure \ref{fig:model_arch}.

\subsection{A volumetric autoencoder to recover gait embedding patterns}\label{sec:autoencoder}

Here, we are interested on capture complex dynamic interactions during locomotion, observed in videos as spatio-temporal textural interactions. From a self-supervised strategy (video-reconstruction task), we implemented a 3D deep autoencoder that projects videos into low-dimensional vectors, learning the complex gait dynamics into a latent space (see the architecture in Figure \ref{fig:model_arch}-a). For doing so, 3D convolutional blocks were implemented, structured hierarchically, with the main purpose to carry out a spatio-temporal reduction while increasing feature descriptions. Formally, a gait sequence $\mathbf{x} \in \mathbb{N}^{f \times h \times w \times c}$, where $f$ denotes the number of temporal frames, $(h \times w)$ are the spatial dimensions, and $c$ is the number of color channels in the video. This sequence is received as input in the convolutional block which is convolved with a kernel $\kappa$ of dimensions ($k_t$, $k_h$, $k_w$), where $k_t$ convolves on the temporal axis and $k_h, k_w$ on the spatial axes. At each level $l$ of processing, we obtain a new volume $\mathbf{x}_l \in \mathbb{Z}^{f/2^l \times h/2^l \times w/2^l \times 2^{l}c}$ that represents a bank of spatio-temporal feature maps. Each of these volumetric features are dedicated to stand out relevant gait patterns in a $\mathbf{z}_G$ reduced projection, that summarizes the receptive fields captured representations.

The resultant embedding vector $\mathbf{z}_G$ encodes principal dynamic non-linear correlations, which are necessary to achieve a video reconstruction $\mathbf{x'}$. In this study, the validated datasets are recorded from a relative static background, so, the major dependencies to achieve an effective reconstruction lies in temporal and dynamic information expressed during the gait. Here, we adopt $\mathbf{z}_{G}$ as a digital gait biomarker that, among others, allows to study motion abnormalities associated to the Parkinson disease. 

To complete end-to-end learning, 3D transposed convolutional blocks were implemented as decoder, positioned in a symmetrical configuration regarding the encoder levels, and upsampling spatio-temporal dimensions to recover original video-sequence. Formally, having the embedded feature vector $\mathbf{z}_{G} \in \mathbb{Z}^{n}$ with $n$ coded features, we obtain $\mathbf{x'}_l \in \mathbb{Z}^{2^{l}f \times 2^{l}h \times 2^{l}w \times c/2^l}$ volumes from transpose convolutional blocks until obtaining a video reconstruction $\mathbf{x'} \in \mathbb{N}^{f \times h \times w \times c}$. The quality of reconstruction is key to guarantee the deep representation learning in the autoencoder part of generator. To do this, an $L_1$ loss is implemented between $\mathbf{x}$ and $\mathbf{x'}$ and its named contextual loss: $L_{con} = \left \| \mathbf{x} - \mathbf{x'} \right \|_1$.

\subsection{Auxiliary task to discriminate reconstructions}\label{sec:discriminator}

From a generative learning, the capability of the deep representations to code locomotion patterns may be expressed in the quality of video reconstructions $\mathbf{x'}$. Hence, we hypothesize that embedding descriptors $\mathbf{z}_G$ that properly reproduce videos $\mathbf{x'}$ should encode sufficient kinematic information of trained class, allowing to discriminate among locomotion populations, \textit{i.e.} between control and Parkinson samples. 

To measure this reconstruction capability, an auxiliary task is here introduced to receive tuples with original and reconstructed videos $\left ( \mathbf{x}, \mathbf{x'} \right )$, and output a discriminatory decision $\mathbf{y} = \left \{ y, y' \right \}$, regarding video source. In such case, $y$ corresponds to the label for real videos ($y=0$), while $y'$ as labels for embeddings from reconstructed sequences ($y=1$). For doing so, we implement an adversarial $L_2$ loss, expressed as: $L_{adv} = \left \|  \mathbf{z}_{D} - \mathbf{z}_{D}' \right \|_2$. In such case, for large differences between $\left ( \mathbf{z}_{D}, \mathbf{z}_{D}' \right )$ it will be a significant error that will be propagated to the generator. It should be noted that such minimization rule optimizes only the generator. Then discriminator is only minimized following a typical equally weighted cross-entropy rule, as: $L_{disc} = \left ( \log(y) + \log(1 - y') \right ) / 2$.  

The auxiliary task to monitor video reconstruction is implemented from a discriminatory convolutional net that follows the same structure that encoder in Figure \ref{fig:model_arch}-a, which halves the spatio-temporal dimension while increases the features and finally dense layer determines its realness level (see in Figure \ref{fig:model_arch}-c.). Interestingly, from such deep convolutional representation the input videos are projected to an embedding vector $\mathbf{z}_{D} \in \mathbb{Z}^{m}$ with $m$ coded features, which thereafter may be used as latent vectors descriptors that also encode motion and realness information.

\subsection{A Digital gait biomarker from anomaly embeddings}\label{sec:embeddings}

To guarantee an optimal coding into low-dimensional embeddings, the reconstructed video $\mathbf{x'}$ is mapped to an additional encoder projecting representation basis in a $\mathbf{z'}_G$ embedding. Hence, we can adopt a new digital kinematic descriptor by considering embedding vector differences between ($\mathbf{z}_G, \mathbf{z'}_G$). In such sense, $\mathbf{z}_G$ and $\mathbf{z'}_G$ must be similar, and lead to $\mathbf{x}$ and $\mathbf{x'}$ to be equal which helps in generator's generalization, following an encoder $L_2$ loss:  $L_{enc} = \left \| \mathbf{z}_{G} - \mathbf{z}_{G}' \right \|_2$. For instance, large difference between $\mathbf{z}_G, \mathbf{z'}_G$ may suggest a new motion class, regarding the original distribution of training. 

From such approximation, we can model a scheme of one-class learning (in this case, anomaly learning) over the video distributions from the low-embedding differences observations. This scheme learns to model the data distribution without any label constraint. Furthermore, if we train the architecture only with videos of a control population ($\mathbf{c}$), we can define a discriminatory problem from the reconstruction, by inducing: $\left \|  \mathbf{z}_{G} - \mathbf{z'}_{G} \right \|_2 \leq \tau \to \mathbf{c} \wedge \left \|  \mathbf{z}_{G} - \mathbf{z'}_{G} \right \|_2 > \tau \to \mathbf{p}$, where $\mathbf{p}$ is a label imposed to a video with a significant error reconstruction and representing Parkinson population.

\subsection{Statistical validation setup}\label{sec:stats}

The video samples are high-dimensional motor observations that can be projected into a low-dimensional embedding space, through the proposed model. The following proposition is used: \\

\begin{remark}
    Given any deep learning model \(F\) with parameters \( \theta \), and any input \( x_{\ell}^{\left ( i \right )} \) of class \( \left ( i \right ) \), belonging to an arbitrary distribution \( \Psi^{\left ( i \right )} \left ( \mathbf{X}^{\left ( i \right )}; \mu^{\left ( i \right )}, \sigma^{\left ( i \right )} \right ) \) where \( x_{\ell}^{\left ( i \right )} \in \mathbf{X}^{\left ( i \right )} \), then the original distribution have the same behaviour and form affecting only the values of \( \mu^{\left ( i \right )} \) and \( \sigma^{\left ( i \right )} \) by applying the model \( F_{\theta} \):
    \begin{equation*}
        \begin{aligned}
            F_{\theta} \left ( x_{\ell}^{\left ( i \right )} \right ) & = y_{\ell}^{\left ( i \right )} \\
            F_{\theta} \left ( \mu^{\left ( i \right )} \right ) & = \hat{\mu}^{\left ( i \right )} \\
            F_{\theta} \left ( \sigma^{\left ( i \right )} \right ) & = \hat{\sigma}^{\left ( i \right )} \Rightarrow \\
            \Psi^{\left ( i \right )} \left ( x_{\ell}^{\left ( i \right )}; \mu^{\left ( i \right )}, \sigma^{\left ( i \right )} \right ) & = \Psi^{\left ( i \right )} \left ( y_{\ell}^{\left ( i \right )}; \hat{\mu}^{\left ( i \right )}, \hat{\sigma}^{\left ( i \right )} \right )
        \end{aligned}
    \end{equation*}
\end{remark}

The above proposition consider the model as non-linear transformation operator applied over the global mean ($\mu^{\left ( i \right )}$) and standard deviation ($\sigma^{\left ( i \right )}$) of class $\left ( i \right )$. From this assumption we can measure statistical properties over high-dimensional space and explore properties as the generalization of the modeling.

The one-class learning scheme can be validated following standard metrics into binary projection $\hat{y} \in \left \{ \mathbf{c}, \mathbf{p} \right \}$. For a particular threshold $\tau$ we can recover metrics such as the accuracy, precision and recall. Also, ROC-AUC  (the Area Under the Curve) can estimate a performance by iterating over different $\tau$ values. However, these metrics only quantify the separation between classes but doesn't exploit the data distribution properties such as shape and variance among classes \cite{demvsar2006statistical,luengo2009study}. To robustly characterize a Parkinson digital biomarker is then demanding to explore more robust statistical alternatives that evidence the generalization of embedding descriptors and estimate the performance for new samples. In fact, we hypothesize that Parkinson and control distributions, observed from an embedding representation, should remain with equal properties from training and test samples. To address such assumption, in this work is explored two statistical properties to validate the shape and variance of motor population distributions: 

\subsubsection{Shape analysis from ChiSquare}\label{sec:shapeness}

Here, we quantify the ``shapenes'' focused in having equally distributions. Particularly, this analysis is carried out for two independent groups $\langle \mathbf{k} \rangle, \langle \mathbf{u} \rangle$ with cardinality $\lvert \mathbf{x}_{\langle \mathbf{k} \rangle}^{(\mathbf{i})}\rvert, \lvert \mathbf{x}_{\langle \mathbf{u} \rangle}^{(\mathbf{j})} \rvert$ of classes $(\mathbf{i}), (\mathbf{j})$. For instance, this metric should be calculated between the clases of parkinson and control using the groups of train, validation and test (\textit{e.g.} compare $\mathbf{x}_{\langle \mathbf{val} \rangle}^{(\mathbf{control})}$ against $\mathbf{x}_{\langle \mathbf{test} \rangle}^{(\mathbf{parkinson})}$). Then, using the ChiSquare test ($\chi^2$) between $\mathbf{x}_{\langle \mathbf{k} \rangle}^{(\mathbf{i})}$ and $\mathbf{x}_{\langle \mathbf{u} \rangle}^{(\mathbf{j})}$, because is a direct comparison for each element as a quadratic relative error: 

\begin{equation}
    \label{eq:chi_general}
    \chi^2_{\langle \mathbf{k} \rangle \to \langle \mathbf{u} \rangle} = \sum_{\ell} \frac{ ( \mathbf{x}_{\ell}^{\langle \mathbf{k} \rangle} - \mathbf{x}_{\ell}^{\langle \mathbf{u} \rangle} )^2 }{\mathbf{x}_{\ell}^{\langle \mathbf{u} \rangle}}
\end{equation}

From this rule, it should be noted that both groups must have the same cardinality ($\lvert \mathbf{x}_{\langle \mathbf{k} \rangle} \rvert = \lvert \mathbf{x}_{\langle \mathbf{u} \rangle} \rvert$) and the respective groups sorting determines the direction of comparison (\textit{i.e.} the direction goes from group $\langle \mathbf{k} \rangle$ to have the same distribution as $\langle \mathbf{u} \rangle$). To address these issues we repeat the lower group in each of its elements without adding new unknown data to preserve its mean and standard deviation, and secondly, we evaluate both directions to match the similarity when $\chi^2_{\langle \mathbf{k} \rangle \to \langle \mathbf{u} \rangle}$ and $\chi^2_{\langle \mathbf{u} \rangle \to \langle \mathbf{k} \rangle}$.

The value $\chi^2_{\langle \mathbf{k} \rangle \to \langle \mathbf{u} \rangle}$ reject the null hypothesis of equal distributions when $\chi^2 > \chi^2_{\alpha, \lvert \mathbf{x}_{\langle \mathbf{u} \rangle} \rvert - 1}$ where $\chi^2_{\alpha, \lvert \mathbf{x}_{\langle \mathbf{u} \rangle} \rvert - 1}$ is the upper critical value of Chi Square distribution with $\lvert \mathbf{x}_{\langle \mathbf{u} \rangle} \rvert - 1$ degrees of freedom at a significance level of $\alpha$ (commonly $5\%$). Then we define the shapeness value as: 

\begin{multline}
    \label{eq:shape_groups}
        Sh ( \mathbf{x}_{\langle \mathbf{k} \rangle}^{(\mathbf{i})}, \mathbf{x}_{\langle \mathbf{u} \rangle}^{(\mathbf{j})} ) = \\
        \begin{dcases}
            \frac{ \chi^2_{\langle \mathbf{k} \rangle \to \langle \mathbf{u} \rangle} + \chi^2_{\langle \mathbf{u} \rangle \to \langle \mathbf{k} \rangle} }{2} & i = j \wedge k \neq u \\
            0 & i = j \wedge k = u \\
            \frac{2 - ( \chi^2_{\langle \mathbf{k} \rangle \to \langle \mathbf{u} \rangle} + \chi^2_{\langle \mathbf{u} \rangle \to \langle \mathbf{k} \rangle} ) )}{2} & i \neq j
        \end{dcases}
\end{multline}

The objective of ``shapenes'' is to evaluate the equality between distributions of the same classes and inequality between distributions of different classes. For instance, when the groups $\mathbf{x}_{\langle \mathbf{val} \rangle}^{(\mathbf{control})}$ and $\mathbf{x}_{\langle \mathbf{test} \rangle}^{(\mathbf{parkinson})}$ must have different distributions so the equation \ref{eq:shape_groups} will be 1 if both distributions are different. On the other hand, the groups $\mathbf{x}_{\langle \mathbf{train} \rangle}^{(\mathbf{control})}$ against $\mathbf{x}_{\langle \mathbf{test} \rangle}^{(\mathbf{control})}$ must have the same distributions, then the ``shapeness'' value must be also 1 if both distributions are equal. The reason of this is based on following the same behaviour as standard metrics like the accuracy or precision.

\subsubsection{Variance analysis from Homoscedasticity}\label{sec:homo}

Here, a equality among variance of data distributions is estimated through homoscedasticity operators. Similarly to ``shapeness'' this metric should be calculated between the clases of parkinson and control for all their groups. Here, it was considered two dispersion metrics regarding the Levene mean ($\Delta_{\ell,\;\mu}^{\langle \mathbf{k} \rangle} = \lvert x_{\ell}^{\langle \mathbf{k} \rangle} - \mu^{\langle \mathbf{k} \rangle} \rvert$), and the Brown-Forsythe median ($\Delta_{\ell,\; med}^{\langle \mathbf{k} \rangle} = \lvert x_{\ell}^{\langle \mathbf{k} \rangle} - med^{\langle \mathbf{k} \rangle} \rvert$). From such dispersion distances, the test statistic $W_{\odot}^{\mathbf{P}}$ between $\mathbf{x}_{\langle \mathbf{k} \rangle}^{(\mathbf{i})}$ and $\mathbf{x}_{\langle \mathbf{u} \rangle}^{(\mathbf{j})}$ can be defined as:

\begin{equation}
    \label{eq:homo_general}
    W_{\odot}^{\mathbf{P}} = \frac{N - \lvert \mathbf{P} \rvert}{\lvert \mathbf{P} \rvert - 1} \sum_{g \in \mathbf{P}} \frac{{(\;\lvert \mathbf{x}_{\langle \mathbf{g} \rangle} \rvert \; (\mu^{\Delta^{\langle \mathbf{g} \rangle} } - \mu^{\Delta})^2\;)}}{{(\;\sum_{\ell \in \langle \mathbf{g} \rangle}{(\Delta_{\ell,\; \odot}^{\langle \mathbf{g} \rangle} - \mu^{\Delta^{\langle \mathbf{g} \rangle} })^2}\;)}}
\end{equation}

where $\odot$ can be the mean or median operation, $P = \{ \mathbf{x}_{\langle \mathbf{k} \rangle}^{(\mathbf{i})}, \mathbf{x}_{\langle \mathbf{u} \rangle}^{(\mathbf{j})}, \cdots \}$ is the union set of every data group from all classes, $\lvert \mathbf{P} \rvert$ is the cardinality of $\mathbf{P}$, $N$ is the sum of each $\lvert \mathbf{x}_{\langle \mathbf{g} \rangle} \rvert$ cardinalities in $P$, $\mu^{\Delta^{\langle \mathbf{g} \rangle}}$ correspond to the mean $\langle \mathbf{g} \rangle$ of $\Delta_{\ell,\; \odot}^{\langle \mathbf{g} \rangle}$ values and $\mu^{\Delta}$ is the overall mean of every $\Delta_{\ell,\; \odot}^{\langle \mathbf{g} \rangle}$ value in $\mathbf{P}$. 

The value $W_{\odot}^{\mathbf{P}}$ rejects the null hypothesis of homocedasticity when $W_{\odot}^{\mathbf{P}} > f_{\alpha, \lvert \mathbf{P} \rvert - 1, N - \lvert \mathbf{P} \rvert}$ where $f_{\alpha, \lvert \mathbf{P} \rvert - 1, N - \lvert \mathbf{P} \rvert}$ is the upper critical value of Fischer distribution with $\lvert \mathbf{P} \rvert - 1$ and $N - \lvert \mathbf{P} \rvert$ degrees of freedom at a significance level of $\alpha$. 

This metric allows to estimate the clustering level for the model and determine if new data samples from another domain are contained in data distributions of control or Parkinson patients. Then, the homoscedasticity value of $\mathbf{x}_{\langle \mathbf{k} \rangle}^{(\mathbf{i})}$ against $\mathbf{x}_{\langle \mathbf{u} \rangle}^{(\mathbf{j})}$ is defined as follow:

\begin{multline}
    \label{eq:homo_groups}
        H ( \mathbf{x}_{\langle \mathbf{k} \rangle}^{(\mathbf{i})}, \mathbf{x}_{\langle \mathbf{u} \rangle}^{(\mathbf{j})} ) = \\
        \begin{dcases}
            \frac{W_{\mu}^{\left \{ \langle \mathbf{k} \rangle , \langle \mathbf{u} \rangle \right \}} +  W_{med}^{\left \{ \langle \mathbf{k} \rangle , \langle \mathbf{u} \rangle \right \}}}{2} & i = j \wedge k \neq u \\
            0 & i = j \wedge k = u \\
            \frac{2 - \left( W_{\mu}^{\left \{ \langle \mathbf{k} \rangle , \langle \mathbf{u} \rangle \right \}} +  W_{med}^{\left \{ \langle \mathbf{k} \rangle , \langle \mathbf{u} \rangle \right \}} \right )}{2} & i \neq j
        \end{dcases}
\end{multline}

Similar to ``shapeness'' the homocedasticity level evaluate in a scale from 0 to 1 if two groups are equal in variance. For instance, the equation \ref{eq:homo_groups} will be 1 when the groups $\mathbf{x}_{\langle \mathbf{val} \rangle}^{(\mathbf{control})}$ and $\mathbf{x}_{\langle \mathbf{test} \rangle}^{(\mathbf{parkinson})}$ different variances. On the other hand, the groups $\mathbf{x}_{\langle \mathbf{train} \rangle}^{(\mathbf{control})}$ and $\mathbf{x}_{\langle \mathbf{test} \rangle}^{(\mathbf{control})}$ will return 1 if both variances are equal. This property is useful when is needed to check if two groups remains in the same distribution range, because two distribution can have the same shape (frequency) but be placed at different numerically domains, quantifying if the model can performance well on any new data domains. Finally, in algorithm \ref{alg:statistical_calculation} is showed the steps to calculate the proposed homoscedasticity and shapeness level for the model and all the pairs of groups of classes.

\begin{algorithm}
    \caption{Calculation of homoscedasticity and shapeness metric for any quantity of data groups with any classes}
    \label{alg:statistical_calculation}
    \begin{algorithmic}
        \Require $C = \left \{  c_0 , c_1, \cdots, c_n \right \}$ \Comment{Classes in dataset}
        \Require $G_{c_i} = \left \{  \mathbf{x}_{\langle 0 \rangle }^{\left ( i \right )} , \mathbf{x}_{\langle 1 \rangle }^{\left ( i \right )}, \cdots, \mathbf{x}_{\langle m_i \rangle}^{\left ( i \right )} \right \} \forall c_i \in C$ \Comment{Partitions per classes}
        \State $h \gets 0$
        \State $s \gets 0$ 
        \For{any pair $( c_i, c_j )$ in $C$}
            \For{any pair $( \mathbf{x}_{\langle k \rangle}^{(i)}, \mathbf{x}_{\langle u \rangle }^{(j)} )$ in $\bigcup ( G_{c_i}, G_{c_j} )$}
                \State $h \gets h + H ( \mathbf{x}_{\langle k \rangle}^{(i)}, \mathbf{x}_{\langle u \rangle}^{(j)} )$ \Comment{$H$ defined in eq. \ref{eq:homo_groups}}
                \State $s \gets s + Sh ( \mathbf{x}_{\langle k \rangle}^{(i)}, \mathbf{x}_{\langle u \rangle}^{(j)} )$ \Comment{$Sh$ defined in eq. \ref{eq:shape_groups}}
            \EndFor
        \EndFor
        \State $N \gets \sum_{i}^{n} \lvert G_{c_i} \rvert$
        \State $d \gets \binom{N}{2}$ \Comment{Combinatory of $N$ in groups of 2}
        \State $h \gets \frac{h}{d}$ \Comment{Homocedasticity level metric}
        \State $s \gets \frac{s}{d}$ \Comment{Shapeness level metric}
    \end{algorithmic}
\end{algorithm}

\section{Experimental setup}\label{sec:setup}
\subsection{Datasets}\label{sec:owner_ds}

In this study 37 patients were recruited having 23 control subjects (average age of $64.7 \pm 13$) and 14 parkinson  subjects (average age of $72.8 \pm 6.8$) populations. The patients were invited to walk (without any markers protocol), developing a natural locomotion gesture. Parkinson participants were evaluated by a physiotherapist (with more than five years of experience) and stratified according to the  H\&Y scale (level 1.0 = 2, level 1.5 = 1, level 2.5 = 5, and level 3.0 = 6 participants). These patients written an informed consent and the total dataset count with the approval of the Ethics Committee of Universidad Industrial de Santander. 

For each recording, a natural gait is developed in around 3 meters, the locomotion was registered 8 times from a sagittal view, following a semi-controlled conditions (a green background). In this study we use a conventional optical camera Nikon D3500, that output sequences at 60 fps with a spatial resolution of 1080p. Each sequence were processed to extract the full gait cycle in 64 frames, spatially resized to 64$\times$64 pixels, and converted to grayscale. Besides, the videos were normalized and cover the whole participant silhouette. To follow one learning class, the proposed approach was trained only with control subjects. In such case, the set of control patients was split in common train, validation and test partitions of 11, 3 and 9 randomly patients, respectively. For parkinson participants, we take for validation and test partitions of 3 and 11 patients randomly selected to complement validation and test control sets. Hence, we balanced validation and test data for standard and statistical validation purposes.

\subsubsection{External dataset validation}\label{sec:ext_ds}

A main interest in this work is to measure the generalization capability of motion patterns from anomaly deep representations. Hence, we are interested in measuring the discrimination performance of the proposed strategy, even for videos captured with external protocols. Then, in this work we only evaluate the proposed approach with a public dataset of walking videos that include knee-osteoarthritis (50 subjects with an average age of 56.7 $\pm$ 12.7), parkinson (16 subjects with an average age of 68.6 $\pm$ 8.3) and control (30 subjects with an average age of 43.7 $\pm$ 9.3) patients \cite{kour2022vision}. The 96 participants have similar characteristics to the owner dataset, but the faces are blurred, the background color changes and there are markers on their bodies. Following the same processing methodology of section \ref{sec:owner_ds}, each sequence was spatially resized to 64$\times$64 pixels, and temporally cropped to 64 frames, and finally normalized and subsampled ensuring a complete gait cycle.

\section{Evaluations and Results}

The proposed strategy was exhaustively validated \textit{w.r.t} the capability to recognize parkinsonian inputs as abnormal class patterns when the architecture is trained only with control patterns and under challenging unbalanced and scarce scenarios. Hence, in the first experiment, the proposed strategy was trained only with control samples from owner dataset, following a video reconstruction pretext task. Hence, encoder ($\left \| \mathbf{z}_{G} - \mathbf{z}_{G}' \right \|_2$), contextual ($\left \| \mathbf{x} - \mathbf{x'} \right \|_1$) and adversarial ($\left \| \mathbf{z}_{D} - \mathbf{z}_{D}' \right \|_2$) errors were recovered as locomotor descriptors of the observed sequences. For classification purposes, these errors were binarized by imposing a threshold value, as: $\tau_{\mathbf{z}_G} = 0.512 $ for encoder, $\tau_{\mathbf{x}} = 0.141 $ for contextual, and $\tau_{\mathbf{z}_D} = 0.813$ for adversarial errors. Table \ref{tab:col_standard_metrics} summarizes the achieved performance of these locomotor descriptors according to standard classification metrics. In general, the proposed strategy reports a remarkable capability to label parkinson patterns as abnormal samples, which are excluded from trained representation. Interestingly, the contextual errors have the highest value among the others to classify between control and parkinson patients, reporting a remarkable 95\% in AUC, with mistakes in only 22 video clips (approximately 2 patients).



\begin{table}[h!]
    \caption{Model performance for encoder, contextual and adversarial losses using standard metrics when the model trains with control patients. Acc, Pre, Rec, Spe, F1 are for accuracy, precision, recall, specificity and f1 score respectively.}
        \begin{tabular}{ccccccc}
            \toprule
            \textbf{Loss} & \textbf{Acc} & \textbf{Pre} & \textbf{Rec} & \textbf{Spe} & \textbf{F1} & \textbf{AUC} \\ \cmidrule(lr){1-1} \cmidrule(lr){2-2} \cmidrule(lr){3-3} \cmidrule(lr){4-4} \cmidrule(lr){5-5} \cmidrule(lr){6-6} \cmidrule(lr){7-7}
            $L_{enc}$ & 67\% & 68\% & 76\% & 54\% & 72\% & 72\% \\ \cmidrule(lr){1-7}
            $\mathbf{L_{con}}$ & \textbf{77\%} & \textbf{99\%} & \textbf{60\%} & \textbf{99\%} & \textbf{75\%} & \textbf{95\%} \\ \cmidrule(lr){1-7}
            $L_{adv}$ & 68\% & 90\% & 49\% & 93\% & 63\% & 76\% \\ 
            \bottomrule
            \label{tab:col_standard_metrics}
        \end{tabular}
\end{table}

\begin{figure}[h!]
    \centering
    \includegraphics[width=0.45\textwidth]{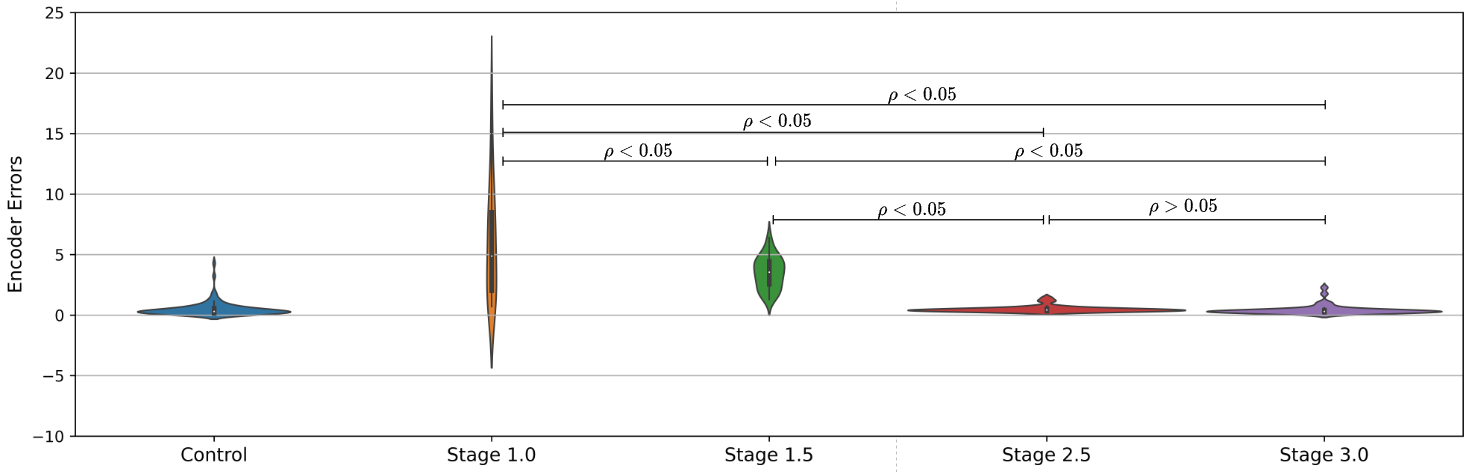}
    \caption{Data distribution given by the proposed model for control and parkinson samples by Hoehn \& Yahr levels.}
    \label{fig:biv_boxplot}
\end{figure}

For robustness validation, we are also interested in the distribution output of predictions, which may suggest the capability of generalization of the model. For doing so, we also validate locomotion descriptors \textit{w.r.t} introduced homoscedasticity and shapeness validation. Table \ref{tab:col_stats_metrics} summarizes the results achieved by each locomotion embedding descriptor, contrasting with the reported results from standard metrics. In such case, the validated metrics suggest that adversarial errors may be overfitted for the trained dataset and the recording conditions, which may be restrictive for generalized architecture in other datasets. Contrary, the encoder descriptor shows evident statistical robustness from variance and shapeness distributions. Furthermore, the encoder losses evidence a clearly separation between the control and parkinson distribution in Figure \ref{fig:biv_boxplot}, where even the proposed model can separate stages of Hoehn \& Yahr with the difference between 2.5 and 3.0 levels where the ChiSquare test shows us that both distributions remains equals meaning that both stages are difficult to model.

\begin{table}[h!]
    \caption{Model performance for encoder, contextual and adversarial losses using the proposed statistical metrics when the model trains with control patients.}
        \begin{tabular}{ccc}
            \toprule
            \textbf{Loss} & \textbf{Homocedasticity} & \textbf{Shapeness} \\ \cmidrule(lr){1-1} \cmidrule(lr){2-2} \cmidrule(lr){3-3}
            \textbf{Encoder} & \textbf{70\%} & \textbf{70\%} \\ \cmidrule(lr){1-3}
            Contextual & 60\% & 40\% \\ \cmidrule(lr){1-3}
            Adversarial & 45\% & 40\% \\ 
            \bottomrule
            \label{tab:col_stats_metrics}
        \end{tabular}
\end{table}

\begin{figure*}[h!]
    \centering
    \includegraphics[width=0.9\textwidth]{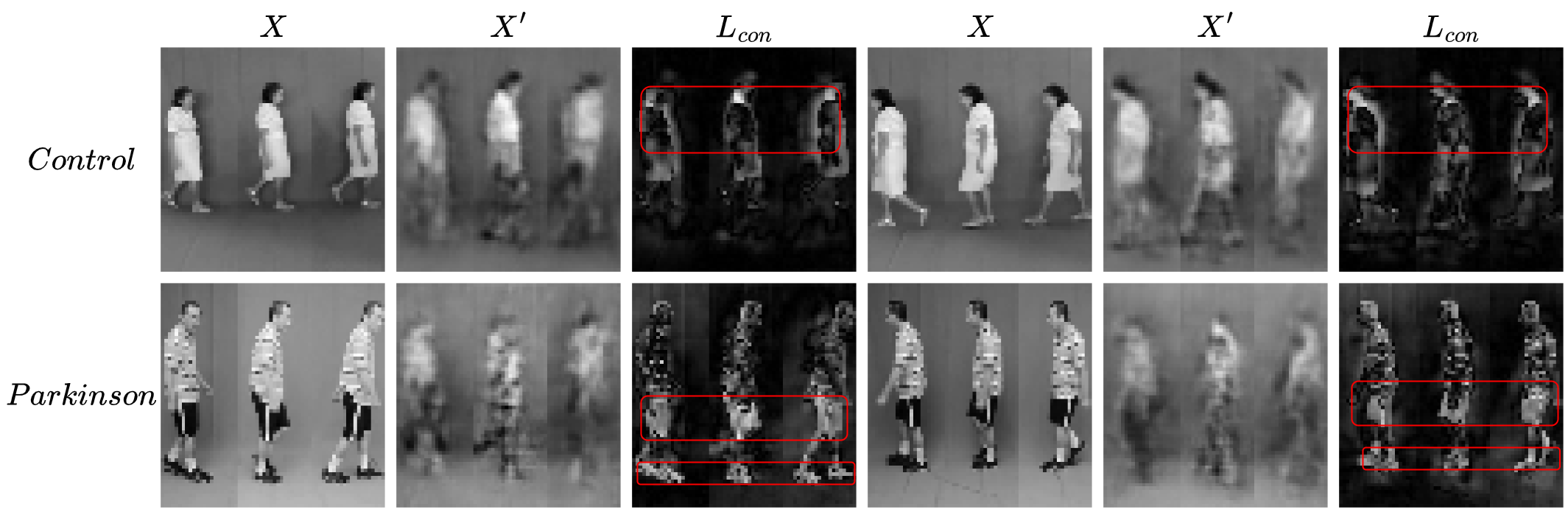}
    \caption{Original, reconstructed and anomaly video of two patients. The first row is a control and second row is a parkinson patient. The 3 first columns are the gait from left to right while the last 3 columns when is right to left.}
    \label{fig:biv_reconstructions}
\end{figure*}

Contrary to discriminative tasks, the proposed approach can generate abnormality videos using the contextual loss ($L_{con}$), as shown in figure \ref{fig:biv_reconstructions}. This figure summarizes a video output with samples at the start, middle and last frames of the sequence. From this abnormality video is possible to determine the key elements which determine why a patient is parkinson or control. 

In the case of a Parkinson patient is common to note more red areas over the legs and feet while control patients have more red areas over the head and arms. In the case of a Parkinson patient is common to have more red areas against the control in the contextual loss, but interestingly these areas are focused on body parts which are abnormal \textit{w.r.t} control locomotion modeled process. Furthermore, these results show that the model is robust against the background and focuses solely on dynamic elements of videos.

To follow with one of the main interests in this work \textit{i.e,} the generalization capability, the proposed strategy was validated with an external public dataset (without any extra training) that include parkinson (16 patients), knee-osteoarthritis (50 patients) and control patients (30 patients) \cite{kour2022vision}. Table \ref{tab:ext_standard_metrics} summarizes the achieved results to discriminate among the three unseen samples, evidencing a notable performance following encoder embedding representation. It should be noted, that Encoder achieves the highest ROC-AUC, reporting an average of 75\%, being the more robust representation, as suggested by statistical homoscedasticity and shapeness validation. The contextual and the adversarial losses have better accuracy, precision and recall, but the specificity suggests that there is not any evidence of correctly classifying control subjects. In such sense, the model label all samples as abnormal from trained representation. In contrast, the encoder element in the network (Figure \ref{fig:model_arch}-a) capture relevant gait patterns to distinguish between control, parkinson and knee-osteoarthritis patients.

\begin{table}[h!]
    \caption{Model performance for encoder, contextual and adversarial losses using the proposed model without retraining and same thresholds as Table \ref{tab:col_standard_metrics}. Acc, Pre, Rec, Spe, F1 are for accuracy, precision, recall, specificity and f1 score respectively.}
        \begin{tabular}{ccccccc}
            \toprule
            \textbf{Loss} & \textbf{Acc} & \textbf{Pre} & \textbf{Rec} & \textbf{Spe} & \textbf{F1} & \textbf{AUC} \\ \cmidrule(lr){1-1} \cmidrule(lr){2-2} \cmidrule(lr){3-3} \cmidrule(lr){4-4} \cmidrule(lr){5-5} \cmidrule(lr){6-6} \cmidrule(lr){7-7}
            $\mathbf{L_{enc}}$ & \textbf{63\%} & \textbf{98\%} & \textbf{58\%} & \textbf{92\%} & \textbf{73\%} & \textbf{75\%} \\ \cmidrule(lr){1-7}
            $L_{con}$ & 87\% & 87\% & 100\% & 0\% & 93\% & 50\% \\ \cmidrule(lr){1-7}
            $L_{adv}$ & 88\% & 89\% & 97\% & 25\% & 93\% & 61\% \\ 
            \bottomrule
            \label{tab:ext_standard_metrics}
        \end{tabular}
\end{table}

\begin{figure}[h]
    \centering
    \includegraphics[width=0.45\textwidth]{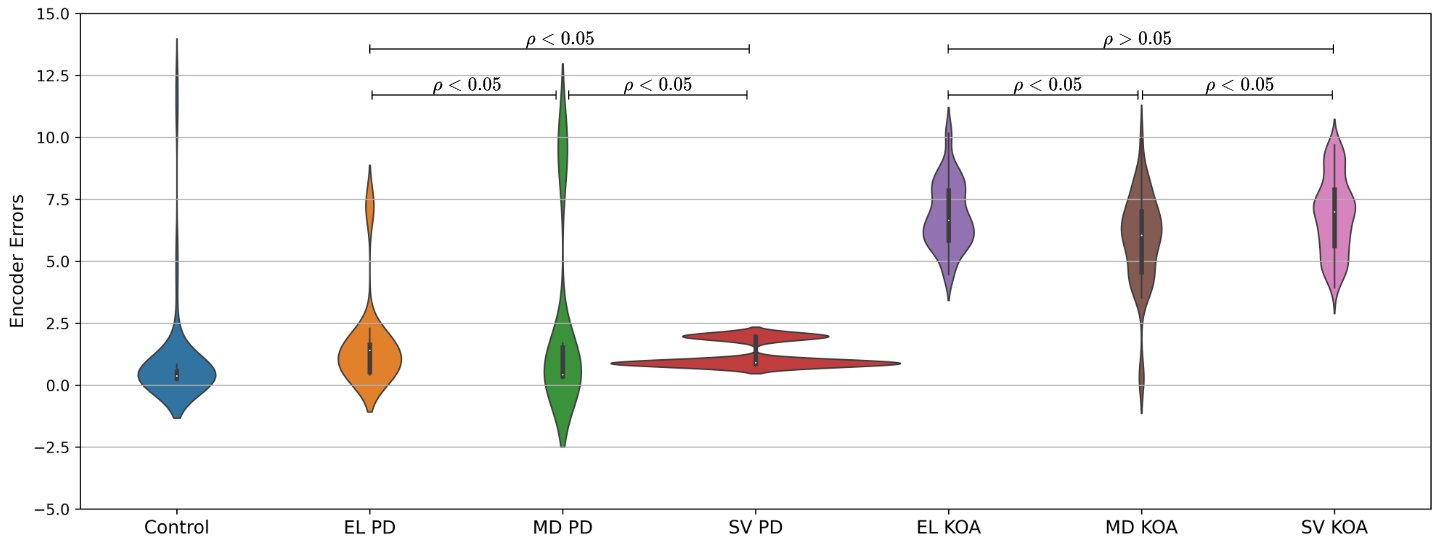}
    \caption{Data distribution given by the proposed model for control, parkinson (PD) and knee-osteoarthritis (KOA) samples by levels where EL is early, MD medium and SV severe.}
    \label{fig:koa_boxplot}
\end{figure}

Along the same line, the external dataset was also validated \textit{w.r.t} homoscedasticity and shapeness metrics. Table \ref{tab:ext_stats_metrics} summarizes the achieved results from the distribution representation of output predictions. As expected, the results enforce the fact that embeddings from the Encoder have much better generalization against the other losses, allowing to discriminate among three different unseen classes. Remarkably, the results suggest that control subjects of the external dataset belong to the trained control set. This fact is relevant because indicates that architecture is principally dedicated to coded locomotor patterns without strict restrictions about captured conditions. To complement such results, output losses from three classes are summarized in violin plots, as illustrated in Figure \ref{fig:koa_boxplot} which shows the separation between the classes of parkinson and knee-osteoarthritis, also, between levels of the diseases, being remarkable the locomotor affectations produced by the patients diagnosed with knee-Osteoarthritis.  

\begin{table}[h!]
    \caption{Model performance for encoder, contextual and adversarial losses using the proposed statistical metrics and model as Table \ref{tab:col_standard_metrics}.}
        \begin{tabular}{ccc}
            \toprule
            \textbf{Loss} & \textbf{Homocedasticity} & \textbf{Shapeness} \\ \cmidrule(lr){1-1} \cmidrule(lr){2-2} \cmidrule(lr){3-3}
            \textbf{Encoder} & \textbf{66.7\%} & \textbf{66.7\%} \\ \cmidrule(lr){1-3}
            Contextual & 83.4\% & 0\% \\ \cmidrule(lr){1-3}
            Adversarial & 16.7\% & 16.7\% \\ 
            \bottomrule
            \label{tab:ext_stats_metrics}
        \end{tabular}
\end{table}


\begin{figure*}[h!]
    \centering
    \includegraphics[width=0.9\textwidth]{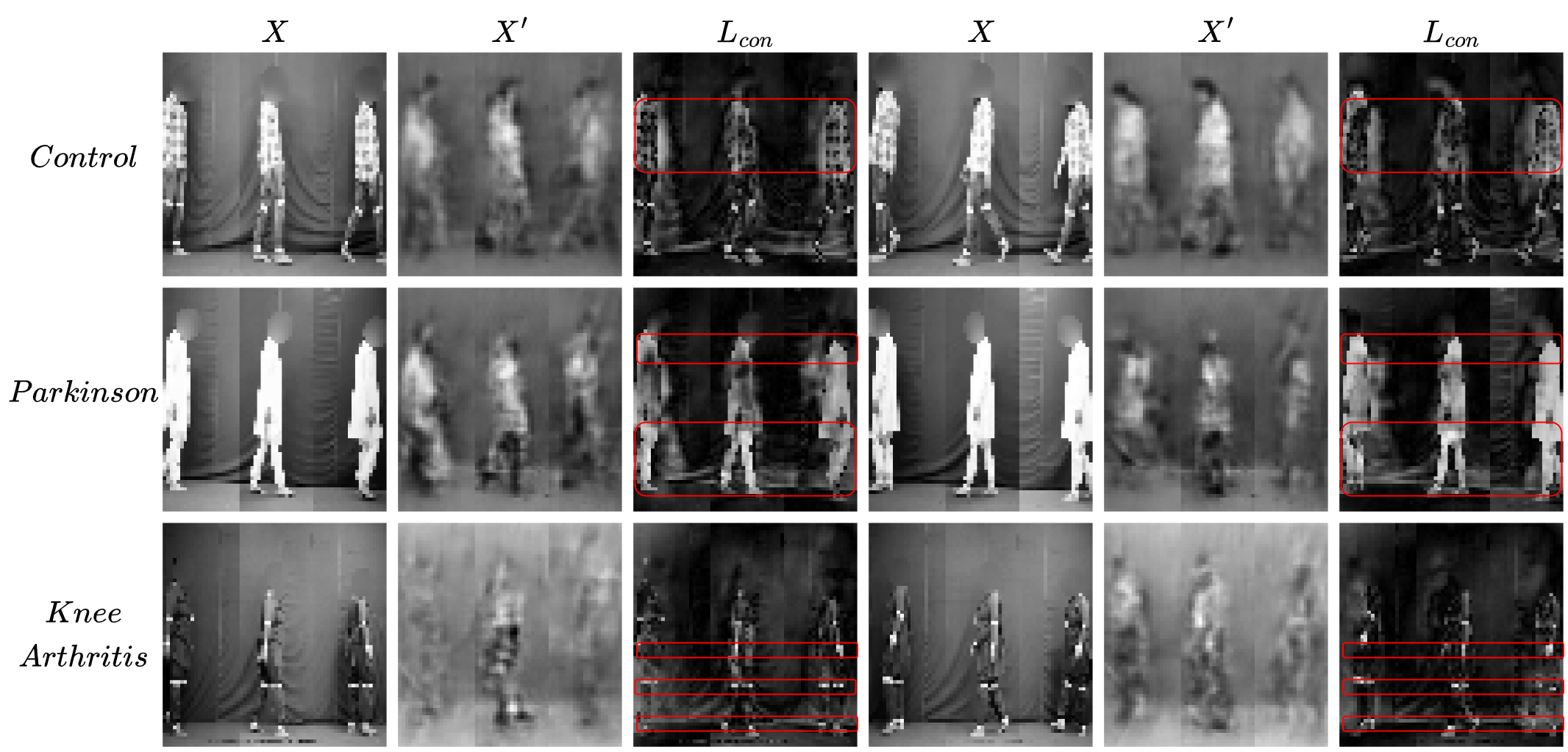}
    \caption{Original, reconstructed and anomaly video of three patients. The first row is a control, second row parkinson and third row is a knee osteoarthritis patient. The 3 first columns are the gait from left to right while the last 3 columns when is right to left.}
    \label{fig:koa_reconstructions}
\end{figure*}

To validate the relation and robustness of the model, we evaluate the abnormality videos for the external dataset using the contextual loss ($L_{con}$), as shown in figure \ref{fig:koa_reconstructions}. This figure evidence a correlation in the abnormal patterns for parkinson patients against the private dataset, where the legs and feet are more highlighted than control patients. In the case of knee-osteoarthritis patients, the abnormality is mainly focused on the knees, hands and feet, which is a common pattern for this disease. Its important to note that the reconstruction videos are noisy due to the bias given by our private dataset, but reconstructions mainly show a control blur gait.

Alternatively, as additional experiment we train using only patients diagnosed with parkinson to force the architecture to extract these abnormal locomotion patterns. In such case, the videos from control subjects are associated with abnormal responses from trained architecture. Table \ref{tab:park_standard_metrics} summarizes the achieved results from standard AUC-ROC and statistical distribution metrics. As expected, from this configuration of the architecture is achieved a lower classification performance because the high variability and complexity to code the disease. In fact, parkinson patients may manifest totally different locomotion affectations at the same stage. For such reason, the architecture has major challenges to discriminate control subjects and therefore lower agreement with ground truth labels. The statistical shapeness metrics confirm such issue achieving scores lower than 50\% and indicating that the model, from such configuration, is not generalizable. In this configuration, it would be demanding a larger amount of parkinson patients to deal with disease variability.

\begin{table}
    \caption{Model performance for encoder, contextual and adversarial losses using standard metrics when the model trains with parkinson patients. Acc, Homo and Shape are for accuracy, homocedasticity and shapeness respectively.}
        \begin{tabular}{ccccc}
            \toprule
            \textbf{Loss} & \textbf{Acc} & \textbf{AUC} & \textbf{Homo} & \textbf{Shape} \\ \cmidrule(lr){1-1} \cmidrule(lr){2-2} \cmidrule(lr){3-3} \cmidrule(lr){4-4} \cmidrule(lr){5-5}
            $L_{enc}$ & 45\% & 54\% & 50\% & 40\% \\ \cmidrule(lr){1-5}
            $L_{con}$ & 54\% & 79\% & 60\% & 40\% \\ \cmidrule(lr){1-5}
            $L_{adv}$ & 41\% & 56\% & 50\% & 40\% \\ 
            \bottomrule
            \label{tab:park_standard_metrics}
        \end{tabular}
\end{table}

\section{Discussion}
This work presented a deep generative scheme, designed under the one-class-learning methodology to model gait locomotion patterns in markerless video sequences. The proposed architecture is trained under the reconstruction video pretext task, being categorical to capture kinematic behaviors without the association of expert diagnosis criteria. From an exhaustive experimental setup, the proposed approach was trained with videos recorded from a control population, while then parkinsonian patterns were associated with anomaly patterns from the design of a discrimination metric that operates from embedding representations. From an owner dataset, the proposed approach achieves an ROC-AUC of 87\%, while for an external dataset without any extra training, the proposed approach achieved an average ROC-AUC of 75\%. 

One of the main issues addressed in this work was to make efforts to train generative architecture with a sufficient generalization capability to capture kinematic patterns without a bias associated to the capture setups. To carefully select such architectures, this study introduced homoscedasticity and shapeness as complementary statistical rules to validate the models. From these metrics was evidenced that encoder embeddings brings major capabilities to generalize models, against the contextual and adversarial losses, achieving in average an 80\% and 70\% for homoscedasticity and shapeness, respectively. Once these metrics defined the best architecture and embedding representation, we confirm the selection by using the external dataset with different capture conditions and even with the study of a new disease class into the population \textit{i.e.,} the Knee-osteoarthritis. Remarkably, the proposed approach generates embeddings with sufficient capabilities to discriminate among different unseen populations as shown in Figure \ref{fig:koa_boxplot}.

In the literature have been declared different efforts to develop computational strategies to discriminate parkinson from control patterns, following markerless and sensor-based observations \cite{balaji2021data, kleanthous2020new, alharthi2020gait, chavez2022vision, guayacan2021visualising, hu2019graph, kour2022classification}. For instance, volumetric architectures have been adjusted from discriminatory rules taking minimization rules associated with expert diagnosis annotations \cite{guayacan2021visualising}. These approaches have reported remarkable results (average an 95\% ROC-AUC with 22 patients). Also, Hu \textit{et. al.} \cite{hu2019graph} proposed an architecture that takes frontal gait views and together with neural graph layers, discriminates the level of freeze in the gait for parkinson patients with an accuracy of 82.5\%. Likewise, Kour \textit{et. al.} \cite{kour2022classification} develops a sensor-based approach to correlate postural relationships with several annotated disease groups (reports an accuracy = 92.4\%, precision = 90.0\% with 50 knee-osteoarthritis, 16 parkinson and 30 control patients). Nonetheless, such schemes are restricted to a specific recording scenario and pose observational configurations. Besides, the minimization of these representations may be biased by label annotations associated with expert diagnostics. Contrary, the proposed approach adjusts the representation using only control video sequences without any expert label intervention during the architecture tunning. In such case, the architecture has major flexibility to code potential hidden relationships associated with locomotor patterns. In fact, the proposed approach was validated with raw video sequences, reported in \cite{kour2022classification}, surpassing precision scores without any additional training to observe such videos. Moreover, the proposed approach uses video sequences instead of representation from key points, that coarsely minimize dynamic complexity during locomotion. 

Recovered generalization metrics scores (homocedasticity  = 80\%, shapeness = 70\% ) suggest that some patients have different statistical distributions, an expected result from variability in control population, as well as, the variability associated to disease parkinson phenotyping. In such sense, it is demanding a large set of training data to capture additional locomotion components, together with a sufficient variability spectrum. Nonetheless, the re-training of the architecture should be supervised from output population distributions to avoid overfitting regarding specific training scenarios. The output reconstruction may also be extended as anomaly maps to evidence in the spatial domain the regions with anomalies, which further may represent some association with the disease to help experts in the correct identification of patient prediction.

\section{Conclusions}
This work presented a deep generative architecture with the capability of discovering anomaly locomotion patterns, convolving entire video sequences into a 3D scheme. Interestingly, a parkinson disease population was projected to the architecture, returning not only outlier rejection but coding a new locomotion distribution with separable patterns \textit{w.r.t} the trained control population. These results evidenced a potential use of this learning and architecture scheme to recover potential digital biomarkers, coded into embedding representations. The proposed approach was validated with standard classification rules but also with statistical measures to validate the capability of generalization. Future works include the validation of proposals among different stages and the use of federated scenarios with different experimental capture setups to test and improve the performance on real scenarios. The model weights and all code used is publicy available in the following repository {\url{https://gitlab.com/bivl2ab/research/2022-edgarrangel_apgn/2023_APN_Anomalous_Parkinson_Network}}.

\backmatter





\bmhead{Acknowledgments}

The authors acknowledge the Ministry of Science Technology and Innovation (MINCIENCIAS), for supporting this research work by the project: "\textit{Caracterización de movimientos anormales del Parkinson desde patrones oculomotores, de marcha y enfoques multimodales basados en visión computational}". Code 92694.

\section*{Declarations}

\bmhead{Acknowledgments}

The authors declare that they have no known competing financial interests or personal relationships that could have appeared to influence the work reported in this paper.

\bmhead{Code availability}

The code that supports the results, graphs and metrics obtained are available online in {\url{https://gitlab.com/bivl2ab/research/2022-edgarrangel_apgn/2023_APN_Anomalous_Parkinson_Network}}.

\bmhead{Availability of data and materials}

The external dataset which supports the results of this study is available in \cite{kour2022vision}. As for the captured dataset will be available on request.








\bibliography{sn-bibliography}

\end{document}